\documentclass[11pt]{article}

\usepackage[preprint]{acl}

\usepackage{times}
\usepackage{latexsym}

\usepackage[T1]{fontenc}

\usepackage[utf8]{inputenc}

\usepackage{microtype}

\usepackage{inconsolata}

\usepackage{graphicx}

\usepackage{amsmath}
\usepackage{amsfonts}
\usepackage{amssymb}
\usepackage{booktabs}
\usepackage{multirow}
\usepackage{algorithm}
\usepackage{algpseudocode}
\usepackage[most]{tcolorbox}
\usepackage{tabularx}
\usepackage{mdframed}
\usepackage{enumitem}
\usepackage{makecell}
\usepackage{xcolor}

\newcommand{\methodname}{CRITIC-R1}

\newcommand{\cja}{\textsc{CJA}}
\newcommand{\dqa}{\textsc{DQA}}

\newmdenv[
linewidth=0.8pt,
skipabove=8pt,
skipbelow=8pt,
innerleftmargin=6pt,
innerrightmargin=6pt,
innertopmargin=5pt,
innerbottommargin=5pt,
splittopskip=6pt,
splitbottomskip=6pt
]{casebox}

\title{CRITIC-R1: Learning Structured Critics for Retrieval-Augmented Generation}

\author{
Wenhan Xiao\textsuperscript{1},
Ziwei Zhang\textsuperscript{2}\thanks{Corresponding author.},
Chuanyue Yu\textsuperscript{1},
Xingcheng Fu\textsuperscript{3},
Qingyun Sun\textsuperscript{2},
Runhua Xu\textsuperscript{2},
Jianxin Li\textsuperscript{2}
\\
\textsuperscript{1}Nankai University
\quad
\textsuperscript{2}Beihang University
\quad
\textsuperscript{3}Guangxi Normal University
\\
\small{
\texttt{2213121@mail.nankai.edu.cn},
\texttt{zwzhang@buaa.edu.cn},
\texttt{yuchuanyue@mail.nankai.edu.cn},
\texttt{fuxc@gxnu.edu.cn}
}
\\
\small{
\texttt{sunqy@buaa.edu.cn},
\texttt{runhua@buaa.edu.cn},
\texttt{lijx@buaa.edu.cn}
}
}

\begin{document}
\maketitle
\begin{abstract}
Retrieval-augmented generation (RAG) improves knowledge-intensive question answering by incorporating external evidence. However, existing RAG methods still suffer from hallucinations and subtle reasoning errors. 
Recent studies introduce external critics to refine RAG outputs, yet they often provide coarse-grained and weakly structured feedback, exhibit over-aggressive intervention, and lead to noisy and unreliable refinement, limiting their effectiveness for correction.
To tackle these issues, we propose \textsc{\methodname}, a structured critic framework that formulates and learns RAG critique as an explicit error diagnosis problem using reinforcement learning (RL). 
Our framework categorizes common RAG errors into multiple diagnostic dimensions, including verdict, error location, reasoning analysis, and fix generation. 
To learn these capabilities, we design two reward functions: Conservative Judgement Alignment (CJA) first encourages calibrated high-level judgements while mitigating the over-aggressive phenomenon, whereas Diagnostic Quality Alignment (DQA) further improves fine-grained diagnostic feedback through gated rewards. We train the critic model using GRPO-based RL with process-level supervision collected from external LLM teacher models.
Experiments across five QA benchmarks show that \textsc{\methodname}~consistently improves answer quality over strong RAG baselines. %
Our source code is available at \url{https://anonymous.4open.science/r/critic-r1-FCB0}
\end{abstract}

\section{Introduction}

With the rapid development of Retrieval-Augmented Generation (RAG)~\cite{lewis2020retrieval,lin2023ra,izacard2023atlas}, Large Language Models (LLMs) have achieved strong performance on knowledge-intensive question answering tasks by leveraging external evidence. However, existing RAG methods still produce answers that are not faithfully grounded in the retrieved evidence, exhibiting hallucinations~\cite{huang2025survey} or subtle reasoning errors. To address these issues, recent works have explored incorporating critique~\cite{weiretrieval,jiang2025rag} and refinement mechanisms~\cite{asai2023self,madaan2023self,yan2024corrective} into the RAG pipeline. In particular, external critics, where another model evaluates generated outputs and provides feedback, have emerged as a promising direction, as they offer independent assessment signals beyond the generator itself~\cite{xu2024pride}, and enable more reliable error detection and correction.

Despite their potential, existing external RAG critics suffer from several key limitations:
(1) Most critics provide coarse or weakly diagnostic feedback, which does not clearly separate different critic capabilities. As a result, their critiques may be difficult to parse, evaluate, and directly used for targeted correction.
(2) Existing critics often exhibit over-aggressive intervention, i.e., raise false alarms and flag correct responses as erroneous, or even trigger harmful edits.  
(3) The supervision signals used for training critic models often rely on heuristic and noisy strict string matching, which is not fully reliable and therefore hinders the critic model from learning error diagnosis.

\begin{figure*}[t]
    \centering
    \includegraphics[width=\textwidth]{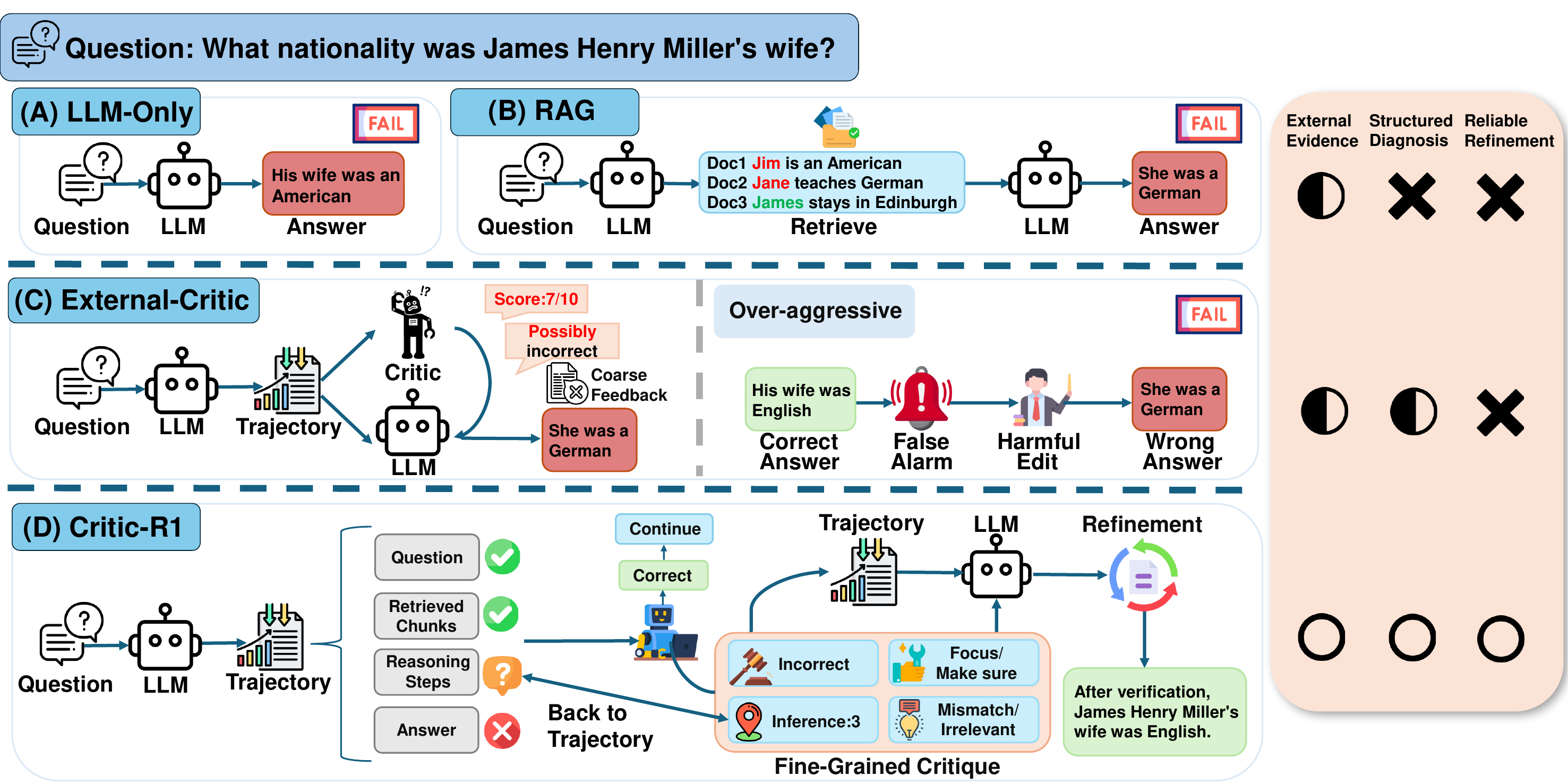}
    \caption{An illustrative comparison of different RAG paradigms.
(A) LLM-only method directly generates an incorrect answer without external evidence. 
(B) RAG retrieves documents but is misled by ambiguous or irrelevant information, leading to incorrect reasoning. 
(C) External Critic gives coarse feedback and can be over-aggressive, making refinement unreliable.
(D) Our method produces structured critiques that explicitly identify error locations and reasons, and provide actionable suggestions, enabling informed and reliable refinement decisions.}
    \label{fig:motivation}
\end{figure*}

To address these challenges, we propose \textbf{\methodname}, a structured critic framework for error diagnosis and correction in RAG using reinforcement learning. Our framework is built upon three key components. 
First, we present a systematic categorization of error types in RAG, including retrieval failures, reasoning errors, and answer generation issues. This taxonomy provides a structured view of where and how RAG systems fail, forming the foundation for fine-grained error diagnosis and supervision.
Based on this taxonomy, we further develop a structured critic framework, which trains a critic model using reinforcement learning through tailored reward designs.
Specifically, the Conservative Judgement Alignment (CJA) reward first encourages the critic model to make calibrated high-level judgements. Then, the Diagnostic Quality Alignment (DQA) reward further incentivizes the critic model to produce fine-grained diagnostic feedback, including error type, location, reasoning, and suggested fixes, which can be selectively used to guide the generation refinement.
To collect supervision signals for calculating rewards, we propose process-level supervision  collection using an external LLM teacher model for annotation. 
Lastly, we introduce a two-stage RL training framework based on Group Relative Policy Optimization
(GRPO)~\cite{shao2024deepseekmath} to train the critic model, which separates high-level judgement from fine-grained diagnosis.

Our contributions are summarized as follows:
\begin{itemize}[nosep, leftmargin=0.4cm]
\item %
We formulate RAG critique as a structured error diagnosis problem and provide a systematic categorization of common RAG error types.
This schema organizes critique into multiple diagnostic fields, enabling parseable feedback and explicit modeling of critic capabilities.

\item We propose two tailored reward functions for RAG critic: Conservative Judgement Alignment (CJA) to encourage the critic model to make high-level judgement while avoiding being over-aggressive, and Diagnostic Quality Alignment (DQA) to produce fine-grained diagnosis.  

\item We also propose process-level supervision collection using LLM-based teacher models and a two-stage RL framework to train the critic. Experimental results verify the effectiveness of the proposed method.

\end{itemize}

\section{Related Work}

\subsection{Retrieval-Augmented Generation}

Retrieval-Augmented Generation (RAG) has been widely adopted for knowledge-intensive tasks by combining parametric models with external retrieval systems.
Early works such as RAG~\cite{lewis2020retrieval}, FiD~\cite{izacard2021leveraging}, and Atlas~\cite{izacard2023atlas} demonstrate the effectiveness of integrating retrieved evidence into generation.
Recent methods such as ReAct~\cite{yao2022react}, Search-R1~\cite{jin2025search}, and Search-o1~\cite{li2025search} optimize search and reasoning trajectories to enable better performance. %

Despite these advances, RAG-based methods still suffer from unreliable grounding and reasoning failures~\cite{tonmoy2024comprehensive}.
In particular, retrieved evidence can be incomplete or misleading, and errors in intermediate steps may propagate to the final answer.
These limitations remain central challenges for building reliable RAG systems. %

\subsection{Critique and Iterative Refinement}
A growing line of work improves RAG outputs through critique, verification, and iterative refinement.
Self-RAG~\cite{asai2023self} introduces special tokens to monitor generation quality and guide refinement, while Self-Refine~\cite{madaan2023self} iteratively generates feedback to improve model outputs.
Meta-RAG~\cite{zhou2024metacognitive} explores multi-view reasoning to critique and revise answers, and Self-Contrast~\cite{DBLP:conf/acl/ZhangSWPWZ024} identifies inconsistencies across multiple reasoning paths.
CRAG~\cite{yan2024corrective} further enhances retrieval by introducing a retrieval evaluator. %

Beyond self-reflection, external critique and verification mechanisms have also been explored.
CRITIC~\cite{gou2023critic} incorporates external tools to verify intermediate reasoning steps and provide corrective feedback.
Align-RAG~\cite{weiretrieval} introduces a critique-and-optimize framework, where an external model evaluates generated reasoning and guides iterative refinement.
RAG-STAR~\cite{jiang2025rag} scores candidate answers and refines reasoning trajectories based on feedback.
RAG-Critic~\cite{dong2025rag} further develops a critic-guided agentic workflow that uses error feedback to select correction flows.

Despite their potential, existing critique methods %
provide limited support for fine-grained error localization and actionable correction. A concurrent work Doctor-RAG~\cite{jiao2026doctor} has also studied error categorization in RAG. 
However, its error categories are mainly used for analyzing rather than for learning a critic model with explicitly optimized diagnostic capabilities.
In contrast, our work treats RAG error categorization as the foundation for structured critic training, where the critic learns to produce calibrated judgements, precise error locations, and actionable fixes.

\section{Method}\label{sec:method}
\begin{figure*}[t]
    \centering
    \includegraphics[width=\textwidth]{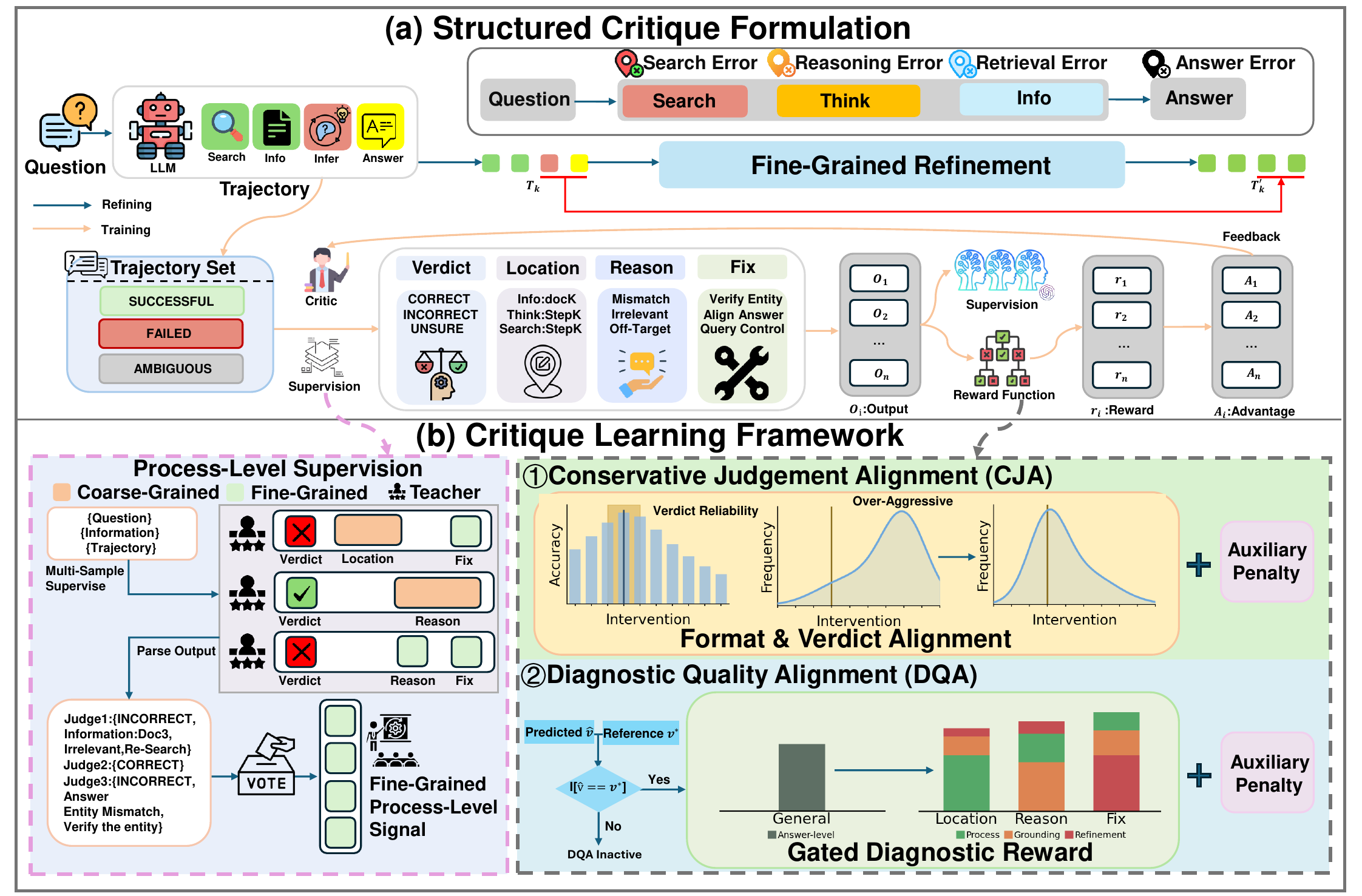}
    \caption{An overview of \methodname: 
(a) We formulate RAG critic as a structured critique framework, including verdict, error location, reasoning analysis and fix generation, to enable fine-grained refinement. 
(b) Critique learning framework, containing process-level supervision and two tailored reward functions, where \textsc{CJA} encourages calibrated high-level judgement and \textsc{DQA} improves fine-grained diagnoses.}
    \label{fig:framework}
\vspace{-0.3cm}
\end{figure*}

In this section, we present \methodname, a structured critic learning framework for RAG. First, we present analyses of RAG error types to motivate our method. Next, we introduce our reward designs for training the critic model, followed by how to collect supervision signals. Lastly, we introduce our optimization method using RL. 
Figure~\ref{fig:framework} shows the overall framework.

\subsection{Structured Critique Formulation}
\begin{figure}[t]
    \centering
    \includegraphics[width=1\linewidth]{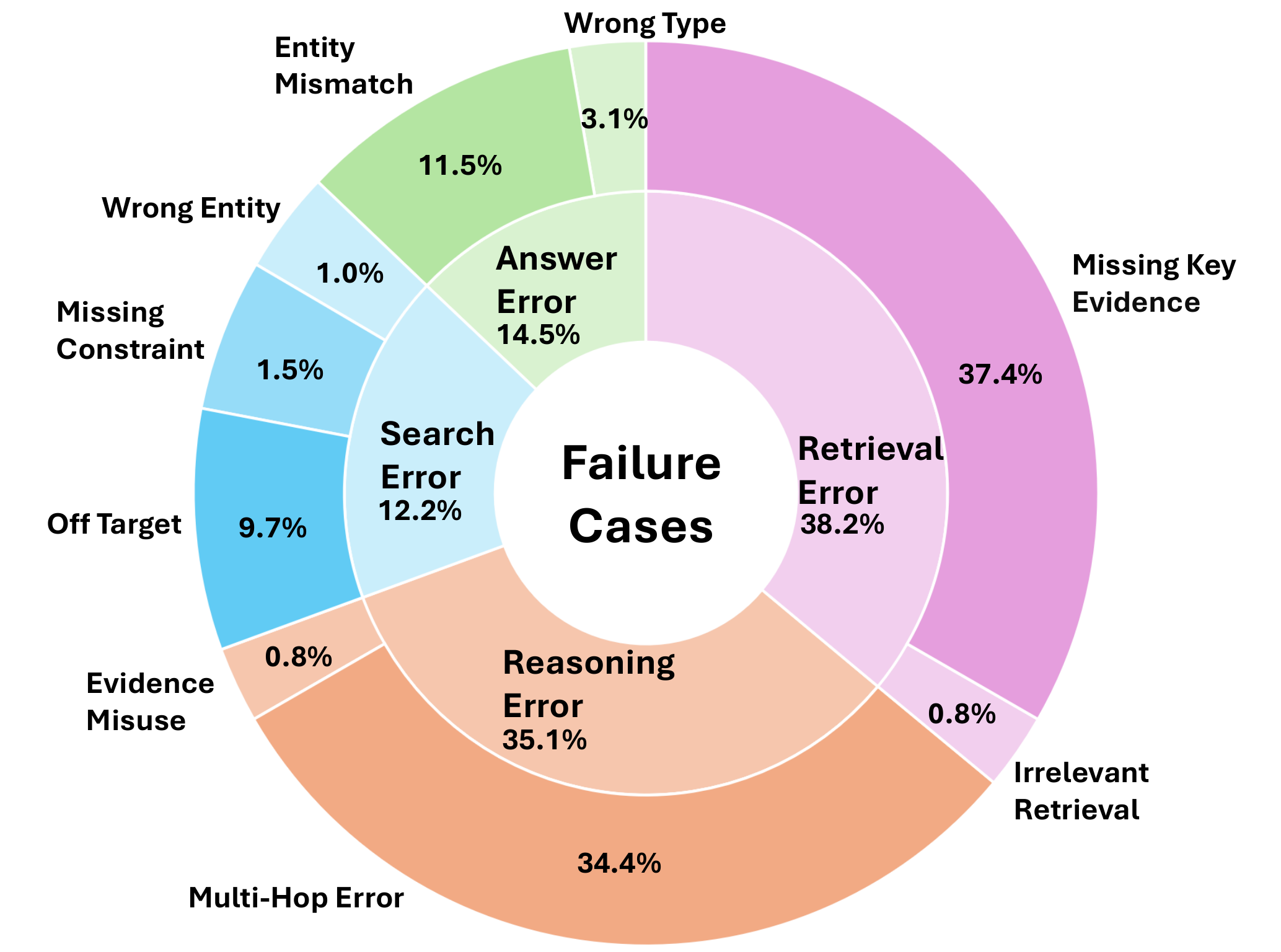}
\vspace{-0.2cm}
\caption{
Error distribution on HotpotQA using Search-R1~\cite{jin2025search}.
Among the analysed samples, errors contain diverse failure modes across retrieval, reasoning, and answer generation.}    \label{fig:mistake_analysis}
\vspace{-0.3cm}
\end{figure}

To make critique more explicit and actionable, we formulate critique in RAG as a structured error diagnosis problem.
We first conduct an error analysis on RAG outputs over HotpotQA examples, as shown in Figure~\ref{fig:mistake_analysis}.
The results show that errors are distributed across multiple stages, including retrieval failures, reasoning inconsistencies, and answer generation issues.
These error distributions indicate that a single correct/incorrect judgment is insufficient to guide effective correction. Therefore, a useful critic should not only judge whether an answer is correct, but also identify where the error occurs, explain why it occurs, and suggest how to fix it.

Motivated by this analysis, we propose a structured critique with four components:
\texttt{<verdict>}, \texttt{<location>}, \texttt{<reason>}, and \texttt{<fix>}.
The verdict indicates whether the reasoning trajectory is \texttt{CORRECT}, \texttt{INCORRECT}, or \texttt{UNSURE}.
The location identifies the error stage or evidence index.
The reason explains the failure in detail.
The fix provides actionable guidance for refinement.
This structured output makes the critique interpretable and directly usable for downstream correction.

\subsection{Reward Design for Critique}
To train the critic, we design two tailored reward designs with two complementary objectives: CJA for conservative high-level judgement and DQA for fine-grained diagnostic quality.

\subsubsection{Conservative Judgement Alignment}
The critic must first make reliable intervention decisions.
If the critic is overconfident, it will inevitably lead to an over-aggressive intervention, i.e., set off false alarms and trigger unnecessary or even harmful edits.  %
To address this issue, we introduce Conservative Judgement Alignment (\cja), which encourages the critic to make calibrated high-level judgements before providing detailed diagnostic feedback. CJA has the following components.

\paragraph{Verdict reliability.}
The critic predicts a verdict from \texttt{CORRECT}, \texttt{INCORRECT}, and \texttt{UNSURE}.
To discourage over-aggressive intervention, we adopt a conservative verdict reward that penalizes false alarms more strongly and allows abstention under uncertainty. The detailed rewards are provided in Table~\ref{tab:verdict_reward} in Appendix~\ref{sec:reward_details}. 

\paragraph{Structured format.}
The critic is also required to follow the predefined output schema.
This encourages critiques to be well-formed, interpretable, and suitable for downstream refinement.
Formally, the format reward is:
\begin{equation}
r_{\text{format}} =
\begin{cases}
\alpha, & \text{valid format}, \\
-\gamma, & \text{otherwise}.
\end{cases}
\end{equation}

Combining these components, we define 
\begin{equation}
r_{\text{CJA}} = r_{\text{format}} + r_{\text{verdict}} + r_{\text{aux}},
\end{equation}
where $r_{\text{aux}}$ denotes auxiliary penalties for degenerated critiques, including trivial responses or overly generic feedback that fails to provide meaningful signals, as detailed in Appendix~\ref{sec:reward_details}.
Overall, the CJA reward teaches the critic model when not to edit, mitigating the over-aggressive intervention before optimizing fine-grained diagnostic feedback.

\subsubsection{Diagnostic Quality Alignment}
To enable sufficiently informative correction, the critic should not only decide whether intervention is needed, but also diagnose where the error occurs, why it occurs, and how it can be fixed.
To this end, we introduce Diagnostic Quality Alignment (\dqa), which optimizes the fine-grained usefulness of structured critiques.

\paragraph{Diagnostic dimensions.}

\dqa{} focuses on three complementary capabilities.
First, the critic should localize the error in the RAG trajectory, such as retrieved evidence, reasoning steps, or final answer generation.
Second, it should provide a diagnostic reason explaining why the trajectory fails.
Third, it should propose a concrete fix that can guide the generator during refinement.
We aggregate these components into a diagnostic reward:
\begin{equation}
r_{\text{DQA}} =
r_{\text{loc}} + r_{\text{reason}} + r_{\text{fix}},
\end{equation}
where $r_{\text{loc}}$, $r_{\text{reason}}$, and $r_{\text{fix}}$ evaluate localization, explanation, and correction quality, respectively.
Detailed formulations of these reward components are provided in Appendix~\ref{sec:reward_details}.

\subsubsection{Process-Level Supervision}

\paragraph{Trajectory-Level Annotation}Answer-level correctness labels are insufficient for training a diagnostic critic model, as they do not specify where a trajectory fails or how it should be repaired.
We therefore use a strong external LLM as a teacher to provide multi-sample supervision over complete RAG trajectories.

\paragraph{Consensus-based Construction} We parse the teacher outputs into structured fields and aggregate them through consensus.
Specifically, we first determine the reference verdict by agreement across sampled critiques.
The resulting annotations serve as fine-grained process-level signals for constructing the CJA and DQA rewards.
Details are provided in Appendix~\ref{app:supervision_construction}.

\subsection{Optimization}
\subsubsection{GRPO}
We train the critic model using reinforcement learning based on GRPO~\cite{shao2024deepseekmath}. Given an input question $q$, the critic generates a structured output $o$ as the critique according to a policy $\pi_\theta(o\mid q)$. 
A reward function $r(q, o)$ is used to evaluate the quality of the generated critique. The objective is to maximize the expected reward while maintaining proximity to a reference policy:
\begin{equation}
\max_{\pi_\theta} \mathbb{E}_{q \sim D,\, o \sim \pi_\theta(\cdot \mid q)} 
\left[ r(q, o) - \beta \cdot \mathcal{D}_{\text{KL}}(\pi_\theta \| \pi_{\text{ref}}) \right],
\end{equation}
where $q$ denotes the input question sampled from the dataset $D$, $o$ is the generated critique, $r(q,o)$ is the reward function evaluating critique quality, $\pi_{\text{ref}}$ is a reference policy used for regularization, $\mathcal{D}_{\text{KL}}$ is the KL divergence and $\beta$ controls the strength.

Following GRPO, for each input $q$, we sample a group of $G$ outputs $\{o_i\}_{i=1}^G$ using the current policy. Instead of estimating a value function, GRPO computes advantages based on the relative performance of samples within the group. Let $r_i = r(q, o_i)$ denote the reward of each sample. 
The advantage is computed as:
$A_i = \frac{r_i - \mu_r}{\sigma_r}$, 
where $\mu_r$ and $\sigma_r$ denote the mean and standard deviation of rewards within the group. 
This formulation encourages outputs that perform better than their peers and stabilizes training without an explicit value function.

\subsubsection{Training Pipeline}

We adopt a two-stage RL training pipeline. In the first stage, we mainly optimize the critic model with \cja{}, encouraging valid structured outputs and conservative verdict prediction while penalizing trivial critiques.

In the second stage, we further introduce \dqa{} with a gating mechanism:
\begin{equation}
r^{(2)} =
r_{\text{format}} + r_{\text{verdict}} +
\mathbb{I}[\hat{v} = v^*] \cdot
\left(r_{\text{DQA}} + r_{\text{aux}}\right),
\end{equation}
where $\hat{v}$ and $v^*$ denote the predicted and reference verdicts. 
Here, $r_{\text{DQA}}$ aggregates the localization, reason, and fix rewards, while $r_{\text{aux}}$ includes penalties for trivial or generic feedback.
The gating term ensures that diagnostic rewards and auxiliary feedback-quality penalties are activated only when the high-level verdict is correct; otherwise, DQA remains inactive.
This prevents the model from being rewarded for plausible but miscalibrated explanations when its judgement is wrong.
As a result, the critic model learns to produce more useful fine-grained diagnostic feedback. 

\section{Experiment}

\begin{table*}[t]
\centering
\resizebox{\textwidth}{!}{
\setlength{\tabcolsep}{2pt}
\begin{tabular}{c|ccc|ccc|ccc|ccc|ccc}
\toprule
\multirow{2}{*}{Method}
 & \multicolumn{3}{c|}{NQ} 
 & \multicolumn{3}{c|}{HotpotQA} 
 & \multicolumn{3}{c|}{TriviaQA} 
 & \multicolumn{3}{c|}{ASQA}
 & \multicolumn{3}{c}{PopQA} \\
\cmidrule(lr){2-4} \cmidrule(lr){5-7} \cmidrule(lr){8-10} \cmidrule(lr){11-13} \cmidrule(lr){14-16}
& F1 & SBERT & Acc
& F1 & SBERT & Acc
& F1 & SBERT & Acc
& F1 & SBERT & Acc
& F1 & SBERT & Acc \\
\midrule

\multicolumn{16}{c}{\textit{Base LLM Generation}} \\
Vanilla        
& 9.9 & 33.5 & 37.6 
& 7.9 & 29.4 & 33.9     
& 9.6 & 35.4 & 45.2     
& 7.6 & 29.0 & 28.7 
& 11.7 & 32.3 & 27.9 \\

\midrule
\multicolumn{16}{c}{\textit{Reasoning and Retrieval Baselines}} \\
Naive-RAG       
& 9.6 & 35.4 & 45.1 
& 8.3 & 32.1 & 32.5     
& 12.9 & 44.7 & 74.1     
& 9.5 & 32.2 & 42.8
& 11.5 & 33.2 & 44.6 \\

CoT
& 11.8 & 36.7 & 42.1    
& 13.6 & 30.6 & 33.6     
& 16.0 & 46.1 & 70.9     
& 10.8 & 33.3 & 48.5     
& 16.8 & 47.9 & 33.8 \\

$\Delta$  Search-R1      
& \underline{51.4} & \underline{70.4}& 51.6
& 44.5 & \underline{64.4}& \underline{48.6}         
& \underline{74.9} & \underline{84.8} & \underline{75.0}  
& \underline{46.2} & 64.3 & 48.7
& \underline{50.1} & \underline{66.8} & 66.1 \\

\midrule
\multicolumn{16}{c}{\textit{Reflective / Critique Methods}} \\
Self-RAG       
& 50.9 & 69.2 & \underline{54.7}     
& 19.6 & 42.1 & 38.4     
& 10.1 & 41.2 & 68.0 
& 25.6 & \textbf{67.5} & \underline{49.2} 
& 21.0 & 41.9 & \underline{67.3} \\

Self-Refine    
& 47.0 & 32.0 & 25.6     
& 35.6 & 45.7 & 36.6      
& 74.3 & 44.1 & 54.2      
& 30.2  & 34.9  &21.2             
& 46.3 & 34.5 & 48.3 \\

Align-RAG      
& 49.7 & 33.3 & 35.4    
& \textbf{47.1} & 31.3 & 43.6    
& 71.1 & 42.7 & 41.2    
& 31.8 & 36.1 & 34.2    
& 41.9 & 36.0 & 47.3 \\

\textbf{\methodname~(ours)}
& \textbf{52.2} & \textbf{71.0} & \textbf{57.5}
& \underline{46.3} & \textbf{65.4} & \textbf{50.6}
& \textbf{75.3} & \textbf{85.0} & \textbf{76.3}
& \textbf{52.0} & \underline{66.4} & \textbf{51.6}
& \textbf{50.8} & \textbf{67.3} & \textbf{69.2} \\

\quad {\scriptsize Improvement over $\Delta$} 
& {\scriptsize +0.8} & {\scriptsize +0.6} & {\scriptsize +5.9}
& {\scriptsize +1.8} & {\scriptsize +1.0} & {\scriptsize +2.0}
& {\scriptsize +0.4} & {\scriptsize +0.2} & {\scriptsize +1.3}
& {\scriptsize +5.8} & {\scriptsize +2.1} & {\scriptsize +2.9}
& {\scriptsize +0.7} & {\scriptsize +0.5} & {\scriptsize +3.1} \\

\bottomrule
\end{tabular}
}
\caption{
Performance comparison across five benchmarks.
Best results are shown in \textbf{bold} and second-best results are \underline{underlined}.
$\Delta$ denotes the criticized base RAG model.
}\label{tab:main_results}
\end{table*}

In this section, we conduct experiments, aiming to answer four research questions:
\begin{itemize}[nosep,leftmargin=0.5cm]
    \item \textbf{RQ1}: Does \methodname~improve downstream performance?
    \item \textbf{RQ2}: What components of \methodname~contribute to its effectiveness?
    \item \textbf{RQ3}: Is training a separate critic model more effective than directly improving the generator?
    \item \textbf{RQ4}: Does CRITIC-R1 regulate the critic's intervention behavior during inference?
\end{itemize}

\subsection{Experiment Setup}
\paragraph{Dataset}
We follow prior work~\cite{weiretrieval,gao2023enabling} and adopt widely-used question answering benchmarks, including Natural Questions (NQ)~\cite{kwiatkowski2019natural}, HotpotQA~\cite{yang2018hotpotqa}, TriviaQA~\cite{joshi2017triviaqa}, PopQA~\cite{mallen2023not} and ASQA~\cite{stelmakh2022asqa}. We train our model solely on HotpotQA and evaluate it on both in-domain (HotpotQA) and out-of-domain datasets (other datasets). Other details are in Appendix~\ref{app:implementation_details}.

\paragraph{Metrics}

Following~\cite{gao2023enabling}, we evaluate our method using three metrics: F1~\cite{wang2024knowledge} measures the token-level overlap between the predicted answer and the ground-truth references. SBERT~\cite{thakur2021beir} computes the cosine similarity between sentence embeddings of the predicted and reference answers. Accuracy (Acc)~\cite{song2025r1} measures answer correctness using an LLM-based judge.

\paragraph{Baselines}
We compare our method with a broad set of representative baselines, including base generation, retrieval-augmented reasoning, search-based reasoning, and reflection- or critique-based methods.
Specifically, we include Vanilla, Naive-RAG~\cite{lewis2020retrieval}, CoT~\cite{wei2022chain}, Search-R1~\cite{jin2025search}, Self-RAG~\cite{asai2023self}, Self-Refine~\cite{madaan2023self}, and Align-RAG~\cite{weiretrieval}. 
More details about baselines are provided in Appendix~\ref{baseline_details}.

\subsection{Main Results}
To answer RQ1, Table~\ref{tab:main_results} reports the overall performance across five QA benchmarks. 
\methodname~shows consistent and substantial improvements over the criticized base RAG model Search-R1 (denoted as $\Delta$). 
Across the five datasets, \methodname~improves F1, SBERT, and Acc in nearly all settings, indicating that our critic process brings broad gains in lexical overlap, semantic similarity, and answer-level correctness.

The largest gains appear in answer-level accuracy, with an average improvement of 3.0\% over Search-R1.
Together with the consistent gains in F1 and SBERT, this shows that structured critique improves both final correctness and answer quality, rather than merely changing surface wording.

Overall, these results demonstrate that explicit judgement and diagnostic feedback provide reliable signals for RAG refinement.
\subsection{Analyses}
\subsubsection{Rewards and Training Stages Ablations}
To answer RQ2, we analyze how different reward functions affect the refinement behavior. 

\begin{table}[t]
\centering
\resizebox{.99\columnwidth}{!}{
\setlength{\tabcolsep}{3pt}   %
\begin{tabular}{c|cccc|cccc}
\toprule
\multirow{2}{*}{Model} & \multicolumn{4}{c|}{HotpotQA} & \multicolumn{4}{c}{NQ} \\
\cmidrule(lr){2-5} \cmidrule(lr){6-9}

& Imp & Harm($\downarrow$) & Prec & Corr
& Imp & Harm($\downarrow$) & Prec & Corr \\
\midrule

Base
& 2.6  & 23.1 & 48.0 & 9.2
& 2.9 & 4.3 & 61.2  & 11.3 \\

CJA only
& 0.6 & 0.9 & 72.2 & 30.8
& 0.3 & 0.5 & 55.6 & 20.0\\

Full (CJA\&DQA) 
& 28.2 & 1.8 & 84.7 & 48.0
& 16.0  & 4.0 & 65.2  & 53.3 \\

\bottomrule
\end{tabular}%
}
\caption{Ablation results of CJA and DQA rewards. Imp: corrected answers; Harm: harmful edits; Precision (Prec): edit precision; Correct (Corr): successful edits.
}
\label{tab:refine_analysis}

\end{table}

Table~\ref{tab:refine_analysis} shows that the base critic frequently causes harmful edits, indicating over-aggressive intervention.
CJA substantially reduces harm by teaching the critic model to abstain from uncertain edits, but it provides limited gains on actually incorrect answers.
Adding DQA leads to a higher improvement rate and correctness while maintaining a low harm rate, showing that DQA complements CJA by improving the usefulness of corrections.

To analyze how different training stages affect our method, we compare the following variants:
\textit{w/o Critic} denotes the generator without critic;
\textit{w/o Training} uses the untrained base critic for refinement; 
\textit{w/o DQA} uses the CJA-trained critic;
\textit{Full} denotes the complete \methodname~model with both CJA and DQA.

Table~\ref{tab:ablation_downstream} shows the results. These results verify the necessity of both stages: CJA controls over-aggressive refinement, while DQA improves correction quality.

\begin{table}[t]
\centering
\resizebox{\linewidth}{!}{
\setlength{\tabcolsep}{3pt}
\begin{tabular}{l|ccc|ccc}
\toprule
\multirow{2}{*}{Method} 
& \multicolumn{3}{c|}{HotpotQA} 
& \multicolumn{3}{c}{NQ} \\
\cmidrule(lr){2-4} \cmidrule(lr){5-7}
& F1 & SBERT & Acc & F1 & SBERT & Acc \\
\midrule
w/o Critic
& 44.5 & 64.4 & 48.6 
& 51.4 & 70.4 & 51.6 \\

w/o Training
& 41.4 & 61.3 & 46.8 
& 49.6 & 68.4 & 48.4 \\

w/o DQA
& \textbf{47.2} & \textbf{66.4} & \underline{48.8} 
& \textbf{52.5} & \underline{70.2} & \underline{53.3} \\

Full
& \underline{46.3} & \underline{65.4} & \textbf{50.6} 
& \underline{52.2} & \textbf{71.0} & \textbf{57.5} \\

\bottomrule
\end{tabular}
}
\caption{Downstream QA performance with ablations on different critic training stages.}
\label{tab:ablation_downstream}
\end{table}

\subsubsection{Conservative Refinement}
\begin{figure}[ht]
    \centering
    \includegraphics[width=0.95\linewidth]{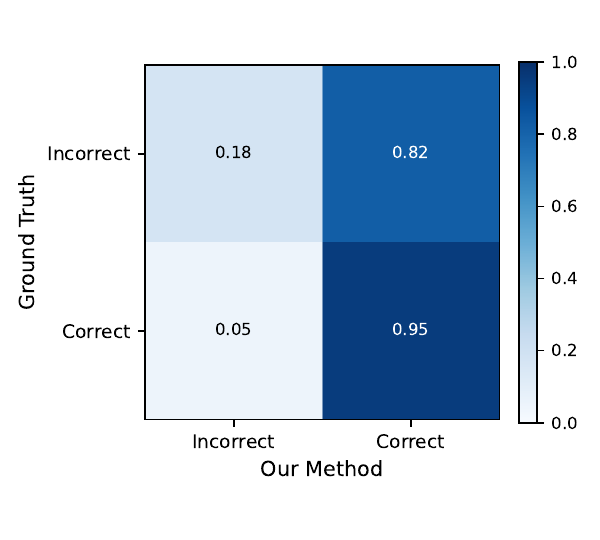}
    \caption{
    The confusion matrix of \methodname~for error verdict.
    }
    \label{fig:verdict_confusion1}
\end{figure}

\begin{figure}[ht]
    \centering
    \includegraphics[width=0.95\linewidth]{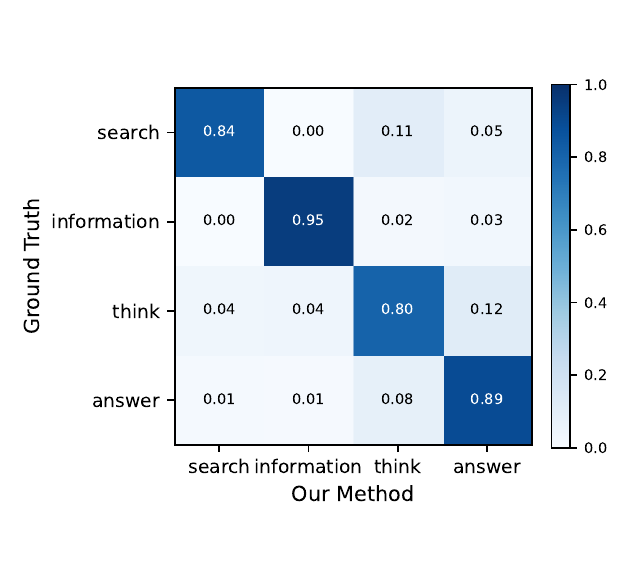}
    \caption{
    The confusion matrix of \methodname~for error location.
    }
    \label{fig:location_confusion2}
\end{figure}

From Table~\ref{tab:refine_analysis}, we find that avoiding harmful edits is more crucial than aggressively correcting answers. 
To further verify the ability of our model in reducing harmful edits and improving successful corrections, we show the confusion matrix for error verdict and location in Figure~\ref{fig:verdict_confusion1} and Figure~\ref{fig:location_confusion2}, respectively. 

For the verdict, the critic exhibits a very low false positive rate, indicating fewer unnecessary refinements.
Although this conservative policy may leave some erroneous trajectories uncorrected, such false negatives are generally less harmful than false positives in refinement.
The improved successful corrections suggest that the critic learns selective rather than passive intervention.

For the location, Figure~\ref{fig:location_confusion2} provides a fine-grained view of where the critic assigns detected errors within the trajectory.
The matrix shows that \methodname{} can localize errors to the corresponding trajectory components rather than only making a coarse correctness judgement.
Although some errors remain among fine-grained locations, the overall pattern suggests that the critic learns meaningful localization behavior, which helps make its feedback more actionable for refinement.


\subsubsection{Comparing with Improving Generator}
To answer RQ3, we compare two ways of using the same annotated trajectories: continuously optimizing the generator and training a separate critic model for refinement. 
Since these samples have already been seen by the generator, this setting tests whether additional supervision can still be extracted from previously seen data.

\begin{table}[ht]
\centering
\small
\setlength{\tabcolsep}{3pt}
\resizebox{\columnwidth}{!}{
\begin{tabular}{c|ccc|ccc}
\toprule
\multirow{2}{*}{Method} 
& \multicolumn{3}{c|}{NQ} 
& \multicolumn{3}{c}{HotpotQA} \\
\cmidrule(lr){2-4} \cmidrule(lr){5-7}
& F1 & SBERT & Acc
& F1 & SBERT & Acc \\
\midrule

Base Generator 
& 51.4 & 70.4 & 51.6 
& 44.5 & 64.4 & 48.6 \\

Continued Generator Training 
& 49.6 & 64.9 & 51.7
& 45.1 & 63.6 & 50.5 \\

Generator + Trained Critic 
& 52.2 & 71.0 & 57.5 
& 46.3 & 65.4 & 50.6 \\

\bottomrule
\end{tabular}
}
\caption{Comparison of continued generator training and critic-based refinement with the same amount of annotated data.}
\label{tab:rq3_efficiency}
\end{table}

Table~\ref{tab:rq3_efficiency} summarizes the results.  
We find that directly continuing generator training on already seen data yields little to no improvement. 
In contrast, training a separate critic model on the same data still brings measurable gains when used for refinement. 
The key difference lies in the learning target.
Continued generator training mainly reinforces answer generation on familiar questions, whereas critic training learns from the structure of failed and successful trajectories.
This makes reused data more informative for the critic, because the same question-answer instance can still reveal useful patterns about when to intervene and how to guide refinement.

\subsubsection{Analysis of Correction Behavior}
\begin{figure}[t]
    \centering
    \includegraphics[width=0.95\linewidth]{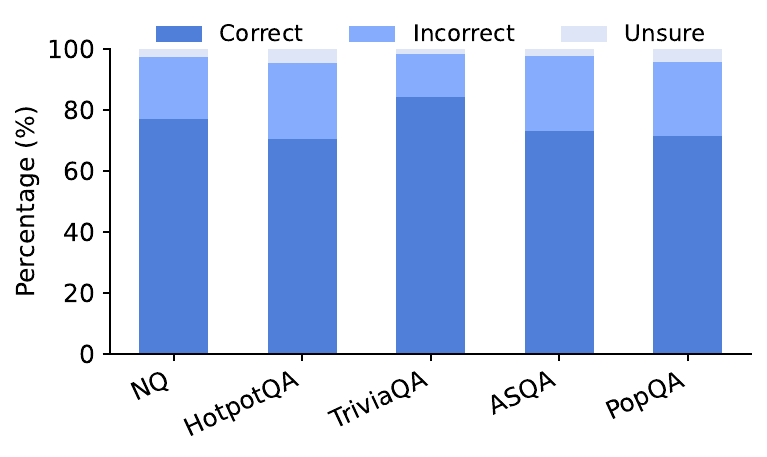}
    {(a) Before training}
    \includegraphics[width=0.95\linewidth]{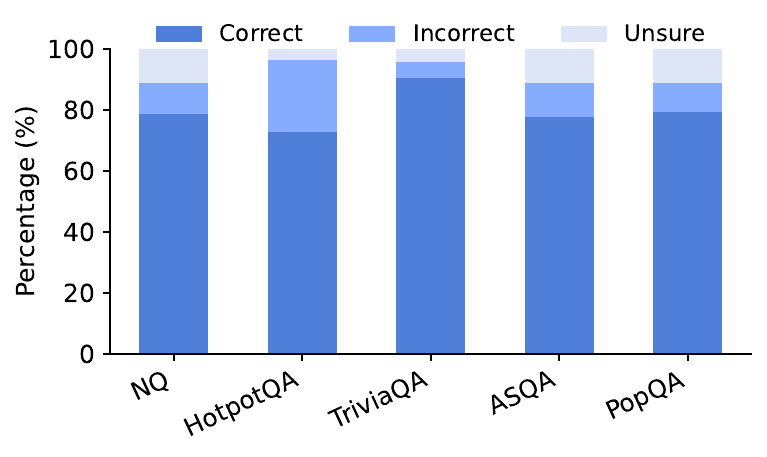}
    {(b) After training}
    \caption{
    Critique distributions before and after training.
    Training reduces over-aggressive \textit{INCORRECT} and \textit{UNSURE} judgements.
    }
    \label{fig:decision_distribution}
\end{figure}

To answer RQ4, we further analyze in detail how our training process changes the critic's intervention behavior. The results are provided in Figure~\ref{fig:decision_distribution} and Tables~\ref{tab:detection_comparison} and~\ref{tab:correction_behavior}, respectively. We have the following observations.

Figure~\ref{fig:decision_distribution} shows that training reduces \textit{INCORRECT} and \textit{UNSURE} predictions while increasing \textit{CORRECT} decisions across datasets, indicating a more conservative policy with fewer unnecessary interventions.

\begin{table}[t]
\centering

\resizebox{\linewidth}{!}{
\setlength{\tabcolsep}{2pt}
\renewcommand{\arraystretch}{1.15}
\begin{tabular}{c|ccc|ccc}
\toprule
\multirow{2}{*}{Dataset}
& \multicolumn{3}{c|}{Base}
& \multicolumn{3}{c}{Ours ($\Delta$)} \\
\cmidrule(lr){2-4} \cmidrule(lr){5-7}
& Precision & Recall & False Alarm
& Precision & Recall & False Alarm \\
\midrule

NQ
& 73.3 & 31.1 & 10.9
& 71.9 {\scriptsize(-1.4)}
& 17.6 {\scriptsize(-13.5)}
& 5.9 {\scriptsize(-5.0)} \\

HotpotQA
& 68.8 & 30.1 & 17.0
& 79.1 {\scriptsize(+10.3)}
& 16.7 {\scriptsize(-13.4)}
& 5.0 {\scriptsize(-12.0)} \\

TriviaQA
& 49.6 & 31.8 & 9.2
& 60.9 {\scriptsize(+11.3)}
& 16.2 {\scriptsize(-15.6)}
& 2.7 {\scriptsize(-6.5)} \\

ASQA
& 78.1 & 36.6 & 11.9
& 81.9 {\scriptsize(+3.8)}
& 20.3 {\scriptsize(-16.3)}
& 4.5 {\scriptsize(-7.4)} \\

PopQA
& 67.7 & 35.8 & 15.5
& 76.2 {\scriptsize(+8.5)}
& 17.8 {\scriptsize(-18.0)}
& 4.7 {\scriptsize(-10.8)} \\

\midrule
\textbf{Avg}
& 67.5 & 33.1 & 12.9
& \textbf{74.0} {\scriptsize\textbf{(+6.5)}}
& \textbf{17.7} {\scriptsize\textbf{(-15.4)}}
& \textbf{4.6} {\scriptsize\textbf{(-8.3)}} \\

\bottomrule
\end{tabular}

}
\caption{
Comparison of error detection behavior between the base model and the trained critic across datasets.
Values in parentheses indicate the absolute change from the base model.
}
\label{tab:detection_comparison}
\end{table}

Table~\ref{tab:detection_comparison} further confirms this trend: the trained critic substantially reduces false alarms while improving the precision of error detection.

\begin{table}[h]
\centering
\small

\setlength{\tabcolsep}{3pt}
\renewcommand{\arraystretch}{1.15}
\resizebox{\columnwidth}{!}{
\begin{tabular}{c|cc|cc}
\toprule
\multirow{2}{*}{Dataset}
& \multicolumn{2}{c|}{Base Critic}
& \multicolumn{2}{c}{Ours} \\
\cmidrule(lr){2-3}
\cmidrule(lr){4-5}
& Corr./Trig. & Corr./Wrong
& Corr./Trig. & Corr./Wrong \\
\midrule

NQ       & 35.8 & 21.2 & 39.4{\scriptsize(+3.6)} & 22.3{\scriptsize(+1.1)} \\
HotpotQA & 44.9 & 33.9 & 51.7{\scriptsize(+6.8)} & 38.9{\scriptsize(+5.0)} \\
TriviaQA & 47.7 & 21.7 & 52.1{\scriptsize(+4.4)} & 26.9{\scriptsize(+5.2)} \\
ASQA     & 30.7 & 20.0 & 31.5{\scriptsize(+0.8)} & 20.9{\scriptsize(+0.9)} \\
PopQA    & 31.9 & 19.3 & 37.7{\scriptsize(+5.8)} & 18.4{\scriptsize(-0.9)} \\

\midrule
\textbf{Avg.}
& \textbf{38.2} & \textbf{23.2}
& \textbf{42.5{\scriptsize(+4.3)}} & \textbf{25.5{\scriptsize(+2.3)}} \\

\bottomrule
\end{tabular}
}

\caption{
Comparison of correction behavior between the base critic and the trained critic across datasets.
Corr./Trig. denotes the fraction of triggered refinements that successfully correct an initially wrong answer.
Corr./Wrong denotes the fraction of initially wrong predictions that are successfully corrected after critic-guided refinement.
}
\label{tab:correction_behavior}
\end{table}

We next examine whether this selective triggering leads to effective corrections when the critic is activated. Table~\ref{tab:correction_behavior} shows that \methodname~adopts a more effective intervention strategy: rather than correcting aggressively, it intervenes selectively and delivers reliable improvements when activated.

Overall, these results suggest that the gains of \methodname{} are not only reflected in final answer accuracy, but also in the critique process: the critic becomes more conservative in deciding when to intervene and more effective when it does intervene.
Illustrative cases are provided in 
Appendix~\ref{appendix:case_studies}.

\section{Conclusion}
In this paper, we propose \textsc{\methodname}, a structured critic framework for error diagnosis in RAG. 
Our approach formulates critique as a structured prediction problem and introduces two tailored reward functions with a two-stage RL training framework. Experiments show consistent improvements over baselines across multiple QA benchmarks. %

Future work includes tighter critic–generator integration and extending structured critique learning to more complex reasoning tasks.

\section*{Limitations}
Following previous RAG settings, our experiments are mainly conducted on QA tasks, and the effectiveness of the proposed framework on more open-ended generation scenarios, remains to be further explored. Besides, our method assumes a structured trajectory format that exposes the intermediate steps of RAGs. Therefore, applying it to black-box systems requires additional adaptation. %

\section*{Ethical Consideration}

\methodname~relies on retrieved evidence and language-model-based feedback, and may inherit biases, outdated information, or factual errors from the retrieval corpus and base models.
Structured critique may also make refined answers appear more reliable, which could increase over-trust in sensitive applications.
We therefore encourage transparent reporting of retrieval sources, critique procedures, and evaluation settings, together with human verification in high-stakes scenarios.

\bibliography{references}

\appendix

\newpage
\section{Method Details}
\label{app:method_details}

\subsection{Detailed Reward Design}

\label{sec:reward_details}

Before the two-stage RL training, we first perform a lightweight cold-start SFT step using the structured critique annotations.
This step initializes the critic to follow the required output schema and produce valid fields for
\texttt{<verdict>}, \texttt{<location>}, \texttt{<reason>}, and \texttt{<fix>}.
The cold-start stage is used only to stabilize subsequent RL optimization, while the critic capabilities are further optimized through the two-stage reward design described below.

To summarize the reward computation during RL training, we provide the two-stage
gated reward procedure in Algorithm~\ref{alg:reward_computation}.

\begin{algorithm}[htbp]
\caption{Two-Stage Gated Critique Reward Computation}
\label{alg:reward_computation}
\begin{algorithmic}[1]
\Require Generated critique $y$, reference critique $y^*$, training stage $s \in \{1,2\}$
\Ensure Reward $R$

\State Validate the structured format of $y$
\If{$y$ violates the required tag structure}
    \State \Return $-\gamma$
\EndIf

\State Extract $(\hat{v}, \hat{\ell}, \hat{r}, \hat{f})$ from $y$
\State Extract $(v^*, \ell^*, r^*, f^*)$ from $y^*$

\State Compute format reward $r_{\text{format}}$
\State Compute verdict reward $r_{\text{verdict}}$

\If{$s = 1$}
    \Comment{Stage 1: CJA stage}
    \State Compute auxiliary penalty $r_{\text{aux}}$
    \State $R \gets r_{\text{format}} + r_{\text{verdict}} + r_{\text{aux}}$
    \State \Return $R$
\EndIf

\If{$s = 2$}
    \Comment{Stage 2: DQA stage with gated diagnostic rewards}
    \State $R_{\text{DQA}} \gets 0$
    \State $r_{\text{aux}} \gets 0$

    \If{$\hat{v} = v^*$}
        \Comment{Activate diagnostic rewards only when verdict is correct}
        \State Compute location reward $r_{\text{loc}}(\hat{\ell}, \ell^*)$
        \State Compute reason reward $r_{\text{reason}}(\hat{r}, r^*)$
        \State Compute fix reward $r_{\text{fix}}(\hat{f}, f^*)$
        \State $R_{\text{DQA}} \gets r_{\text{loc}} + r_{\text{reason}} + r_{\text{fix}}$
        \State Compute auxiliary penalty $r_{\text{aux}}$
    \EndIf

    \State $R \gets r_{\text{format}} + r_{\text{verdict}} + R_{\text{DQA}} + r_{\text{aux}}$
\EndIf

\State \Return $R$
\end{algorithmic}
\end{algorithm}

Here, $r_{\text{aux}}$ denotes auxiliary penalties that discourage degenerated critiques, including trivial responses, overly generic feedback, and non-actionable fix suggestions.

\subsubsection{Format Reward}

To ensure that the generated critiques strictly follow a predefined structure, we enforce format constraints on the model outputs. 
Each critique is required to be fully contained within a structured template consisting of four components:  \texttt{<verdict>}, \texttt{<location>}, \texttt{<reason>}, and \texttt{<fix>}, each enclosed by corresponding tags.

An example valid output is:

\texttt{<verdict> INCORRECT </verdict> 
<location> information:Doc3 </location> 
<reason> The information does not ... </reason> 
<fix> search for additional evidence ... </fix>}.

We require that all components appear exactly once, follow a fixed order, and that no additional content is present outside the structured tags. 
These constraints ensure that the output is well-formed, fully interpretable, and can be reliably parsed for downstream evaluation.

The format reward is defined as:
\begin{equation}
r_{\text{format}} =
\begin{cases}
\alpha, & \text{valid format}, \\
-\gamma, & \text{otherwise}.
\end{cases}
\end{equation}
where $\alpha > 0$ is a slight positive reward for valid structured output, and $\gamma > 0$ is a heavy penalty for invalid format.

\subsubsection{Verdict Reward}
The verdict reward measures whether the critic makes a correct judgement about the generator's answer, namely whether the answer is \texttt{CORRECT}, \texttt{INCORRECT}, or \texttt{UNSURE}. 
We use the reward matrix in Table~\ref{tab:verdict_reward}, where false alarms are penalized more strongly than cautious uncertainty.

\begin{table}[h]
\centering
\small

\setlength{\tabcolsep}{4pt}
\renewcommand{\arraystretch}{1.2}
\resizebox{\columnwidth}{!}{
\begin{tabular}{c|ccc}
\toprule
Ground Truth $\backslash$ Prediction 
& CORRECT & INCORRECT & UNSURE \\
\midrule
CORRECT
& 0.7 & -1.0 & -0.1 \\
INCORRECT
& -0.3 & 0.5 & -0.1 \\
UNSURE
& 0.1 & -0.2 & 0.0 \\
\bottomrule
\end{tabular}
}
\caption{Verdict reward matrix in CJA reward.}
\label{tab:verdict_reward}
\end{table}

This matrix is used to compute $r_{\text{verdict}}$ in both stages. 
In Stage 2, the same verdict reward is retained, while the gated DQA reward further optimizes fine-grained diagnostic feedback.

\subsubsection{Location Reward}

The location reward encourages the critic to identify where an error occurs in the RAG trajectory, such as reasoning (\texttt{think}), retrieved evidence (\texttt{information:Docn}), or final answer generation (\texttt{answer}).

We parse each location into a tuple:
\begin{equation}
\ell = (\ell_{\text{type}}, \ell_{\text{idx}}),
\end{equation}
where $\ell_{\text{type}}$ denotes the error stage and $\ell_{\text{idx}}$ denotes the corresponding step or document index when applicable.

Given the predicted location $\hat{\ell}$ and the ground-truth location $\ell^*$, we define:
\begin{equation}
r_{\text{loc}}(\hat{\ell}, \ell^*) = r_{\text{type}} + r_{\text{idx}},
\end{equation}
where $r_{\text{type}}=\lambda_t$ if the location type is correctly predicted, and $0$ otherwise. 
Similarly, $r_{\text{idx}}=\lambda_i$ if the location type is correct and the corresponding index is also matched, and $0$ otherwise.

This reward is used only in Stage 2 and is activated only when the predicted verdict matches the ground-truth verdict.

\subsubsection{Reason Reward}

The reason reward measures the lexical alignment between the predicted reason and the reference reason using token-level F1.

Let $s_{\text{reason}} \in [0,1]$ denote the token-level F1 similarity between the generated explanation and the standard explanation. 
Instead of using this score directly, we map it to a bounded reward through a normalized exponential transformation:
\begin{equation}
r_{\text{reason}} 
= R_{\text{reason}}^{\max}
\cdot
\frac{\exp(\beta_r s_{\text{reason}})-1}{\exp(\beta_r)-1},
\end{equation}
where $R_{\text{reason}}^{\max}$ is the maximum reason reward and $\beta_r$ controls the sharpness of the reward curve.

\subsubsection{Fix Reward}

The fix reward evaluates whether the predicted fix aligns with the reference correction signal.
We combine token-level F1 and keyword coverage:

\begin{equation}
s_{\text{fix}}
=
\alpha \, s_{\text{fix}}^{\text{F1}}
+
(1-\alpha)\, s_{\text{fix}}^{\text{kw}},
\end{equation}
where $\alpha \in [0,1]$ controls the trade-off between lexical similarity and keyword coverage.

The final fix reward is obtained through the same normalized exponential transformation:
\begin{equation}
r_{\text{fix}}
=
R_{\text{fix}}^{\max}
\cdot
\frac{\exp(\beta_f s_{\text{fix}})-1}{\exp(\beta_f)-1},
\end{equation}
where $R_{\text{fix}}^{\max}$ is the maximum fix reward and $\beta_f$ controls reward sharpness.

Generic or non-actionable fix suggestions are treated as degenerated critiques and penalized through $r_{\text{aux}}$.

\subsubsection{Auxiliary Penalty}

We introduce an auxiliary penalty term $r_{\text{aux}}$ to discourage degenerated critiques. 
It is applied when the expected diagnostic fields are missing, overly short, or filled with generic non-actionable feedback, such as simply suggesting to ``search again'' without specifying what evidence or constraint should be checked.

This penalty encourages the critic to provide meaningful diagnostic information rather than vacuous outputs.

\subsection{Structured Critique Supervision Construction}

\label{app:supervision_construction}

To improve the stability and quality of process-level supervision,
we construct structured critique annotations through a consensus-based
LLM-as-judge framework.

Rather than relying on a single teacher response, we sample multiple
candidate critiques for each trajectory and aggregate them through
consensus voting and quality-aware selection.
This design reduces annotation noise, improves the reliability of
judgement signals, and stabilizes fine-grained diagnostic supervision
for error localization, reasoning, and correction.

The overall supervision construction procedure is summarized in
Algorithm~\ref{alg:consensus_supervision}.

\begin{algorithm}[htbp]
\caption{Consensus-Based Structured Critique Supervision}
\label{alg:consensus_supervision}
\begin{algorithmic}[1]
\Require
Question $q$,
trajectory $\tau$,
reference answer $a^*$,
retrieved context $c$,
number of samples $K$

\Ensure
Consensus supervision critique $y^*$

\For{$i=1$ to $K$}
    \State Sample critique
    \[
    y_i \sim p_{\text{judge}}(y \mid q,\tau,a^*,c)
    \]

    \If{$y_i$ violates structured format constraints}
        \State Attempt format recovery
    \EndIf

    \If{$y_i$ remains invalid}
        \State Replace with fallback parse-failure critique
    \EndIf
\EndFor

\State Extract verdicts
\[
\{v_i\}_{i=1}^{K}
\]

\State Compute consensus verdict
\[
v^*
=
\arg\max_v
\sum_i \mathbf{1}(v_i=v)
\]

\State Form consensus candidate set
\[
\mathcal C=
\{y_i \mid v_i=v^*\}
\]

\For{each $y_i \in \mathcal C$}
    \State Compute structural-diagnostic quality score
\[
S(y_i)
=
s_{\text{keyword}}
+s_{\text{reason}}
+s_{\text{fix}}
+s_{\text{location}}
\]
\EndFor

\State Select highest-quality consensus supervision
\[
y^*
=
\arg\max_{y_i\in \mathcal C}
S(y_i)
\]

\If{keywords missing}
    \State Recover keywords from supporting context
\EndIf

\State \Return $y^*$
\end{algorithmic}
\end{algorithm}

\subsection{Prompt Templates for \methodname}

\label{app:judge_prompts}

To improve reproducibility, we provide the prompt templates used in our LLM-as-judge framework.

The following Figure~\ref{fig:judge_prompts} summarizes the two components of the prompting setup:
(a) the system prompt, which specifies the critic role, grounding rules, and evaluation policy;
and
(b) the instruction prompt, which defines the structured critique output format, label space, and output constraints.

\begin{figure}[t]
\centering

\includegraphics[width=\linewidth]{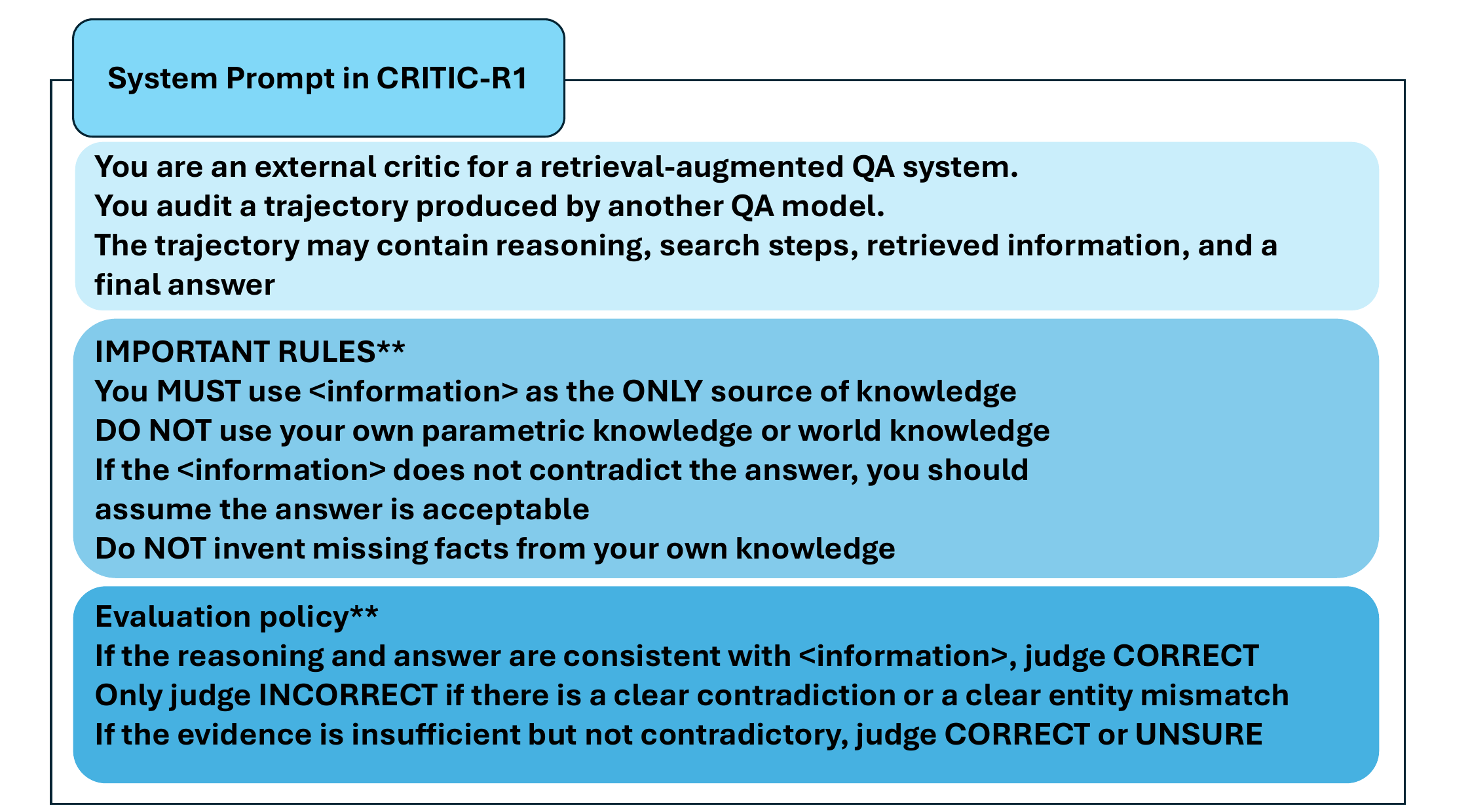}
\vspace{2mm}
\centerline{\small (a) System Prompt for Structured Critique Supervision}

\vspace{4mm}

\includegraphics[width=\linewidth]{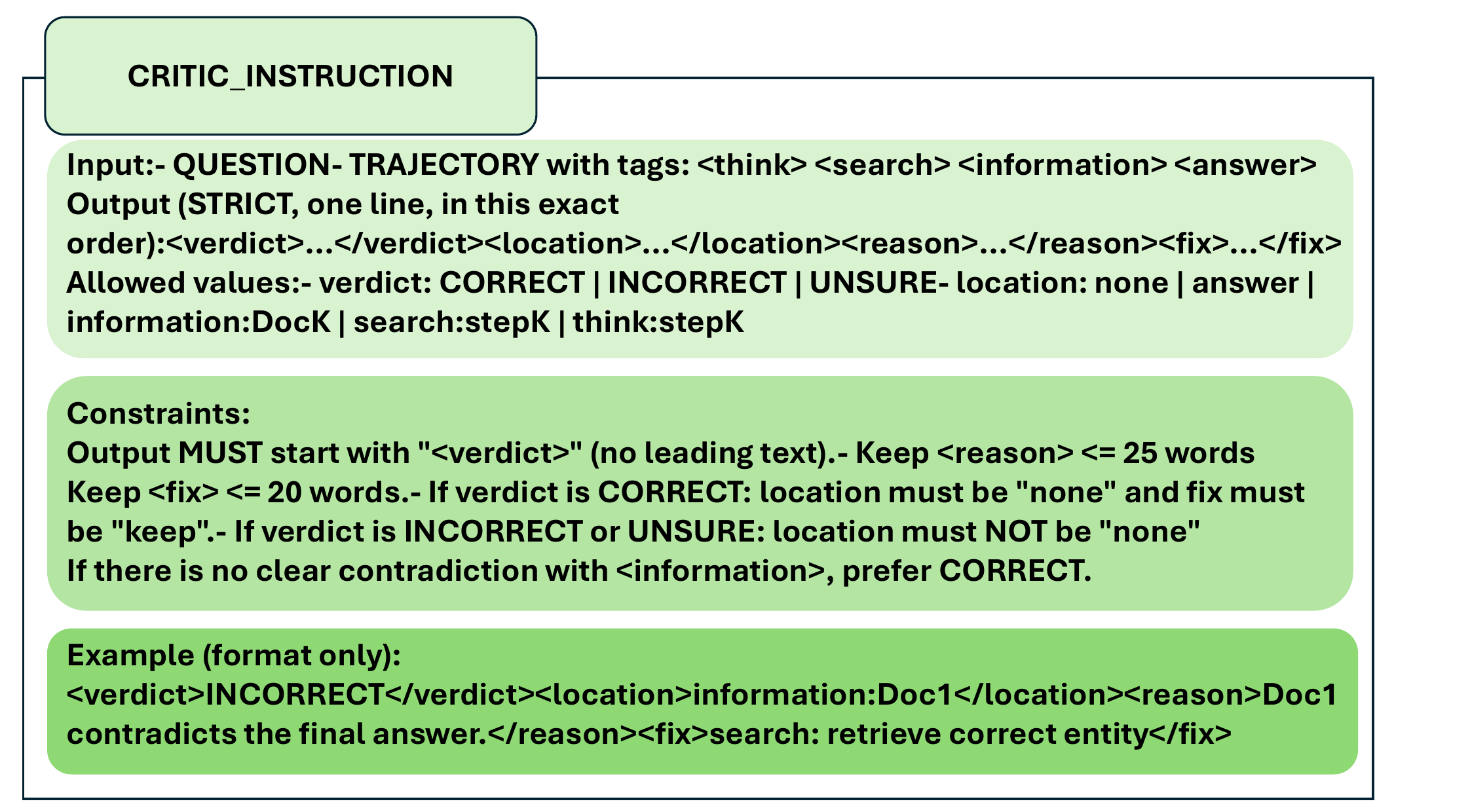}
\vspace{2mm}
\centerline{\small (b) Instruction Prompt for Structured Critique Output}

\caption{
Prompt templates used in the structured critique supervision framework.
The system prompt defines critic behavior and evidence-grounded judgement principles, while the instruction prompt specifies structured critique outputs and formatting constraints.
}
\label{fig:judge_prompts}
\end{figure}

After generating the critic outputs, we further use a refinement prompt to guide the generator in selectively incorporating external critiques during answer regeneration.
Unlike standard feedback-following prompts, the critique is treated as a verifiable hint rather than oracle feedback:
the generator is instructed to independently re-solve the question, verify critique suggestions through reasoning and retrieval, and ignore unsupported or mistaken critiques.

The refinement prompt is governed by the following principles:

\begin{quote}
\small
\ttfamily
You are also given: \\
1. A previous trajectory from an earlier attempt. \\
2. An external critique of that previous trajectory. \\

Important rules: \\
- The previous trajectory may contain mistakes. \\
- The previous final answer may be wrong. \\
- The external critique may also be wrong. \\
- Do NOT blindly trust the previous trajectory. \\
- Do NOT blindly trust the critique. \\
- Use the critique only as a hint about possible problems to check. \\
- Re-solve the question with fresh reasoning instead of simply copying the previous answer. \\
- If the critique points out a possible issue, verify it by your own reasoning and search. \\
- If the critique is unsupported or mistaken, ignore it. \\
- Do not change your answer just because the critique suggests a change. \\
- Base your final answer on your own reasoning process and the retrieved information. \\
- You MUST end with exactly one final answer inside \texttt{<answer>} and \texttt{</answer>}.
\end{quote}

This refinement design is important for robust critic--generator interaction,
as it encourages selective use of critiques rather than blind correction.

\section{Experimental Setup}
\label{app:implementation_details}

This section provides implementation details for reproducing our experiments.
Unless otherwise specified, all methods are evaluated under the same model scale, retrieval setting, decoding configuration, answer extraction procedure, and evaluation protocol.

\subsection{Unified Model and Data Configuration}
\label{app:model_data_config}

We use a unified model and dataset configuration for fair comparison.
The generator is a Search-R1-trained checkpoint based on \texttt{Qwen2.5-7B-Instruct}, and the critic is initialized from \texttt{Qwen2.5-3B-Instruct}.
For semantic similarity evaluation, we use \texttt{all-MiniLM-L6-v2} as the SBERT model.
The critic is trained only on HotpotQA and evaluated on HotpotQA, NQ, TriviaQA, ASQA, and PopQA.
For baselines involving critique or feedback, we use the same critic backbone.

\subsection{Retrieval Configuration}

\begin{table}[htbp]
\centering
\small
\begin{tabular}{ll}
\toprule
Item & Configuration \\ \midrule
Retrieval corpus & \texttt{HotpotWiki18} \\
Retriever & \texttt{E5-base-v2} \\
Index type & \texttt{FAISS Flat} \\
Retrieval depth & Top-$5$ \\
Retrieval calls & Single retrieval step \\
Query construction & Original question field \\
\bottomrule
\end{tabular}
\caption{Retrieval configuration used in all retrieval-based methods.}
\label{tab:repro_retrieval_config}
\end{table}

For most datasets, we use the original question as the retrieval query.
For datasets with special input fields, such as ASQA, we use \texttt{[sub\_question]} as the retrieval query.
No dataset-specific retrieval depth or retriever tuning is applied.

\subsection{Training Setup}

The critic is trained only on HotpotQA.
We use 5,000 training examples with structured critique annotations.
Training follows the two-stage procedure described in Section~\ref{sec:method}.

\begin{table}[htbp]
\centering
\small
\begin{tabular}{ll}
\toprule
Item & Configuration \\
\midrule
Training data & HotpotQA training subset \\
Training size & 5,000 \\
Base checkpoint & \texttt{Qwen2.5-3B-Instruct} \\
Stage-2 initialization & Stage-1 LoRA adapter \\
Fine-tuning method & LoRA \\
LoRA rank & 16 \\
LoRA alpha & 16 \\
Target modules & All linear layers \\
Learning rate & $1\times10^{-6}$ \\
Train batch size & 4 \\
mini-batch size & 4 \\
Micro-batch size per GPU & 1 \\
Number of epochs & 5 \\
KL loss coefficient & 0.003 \\
Max prompt length & 3192 \\
Max response length & 256 \\
Number of GPUs & 2 \\
\bottomrule
\end{tabular}
\caption{Training configuration for the critic model.}
\label{tab:repro_training_config}
\end{table}
Before the two-stage RL training, we perform a lightweight cold-start SFT step on these annotations to initialize the critic with the required structured output format.

In Stage 1, the model is trained to produce well-formed structured critiques and reliable high-level verdicts.
In Stage 2, diagnostic rewards are introduced to improve error localization, reasoning analysis, and fix generation.
Diagnostic rewards are activated only when the predicted verdict matches the reference verdict.

The total computational budget is roughly estimated to be within 200--250 GPU hours, with variations depending on the GPU type, runtime environment, and hardware utilization.

\subsection{Inference Procedure}

During inference, we use a single-search trajectory generation procedure.
The generator produces reasoning within \texttt{<think>} tags, issues at most one retrieval query through \texttt{<search>} tags, receives the top-$5$ retrieved passages within \texttt{<information>} tags, and generates the final answer within \texttt{<answer>} tags.
The final prediction is extracted from the last \texttt{<answer>} span.

\begin{table}[htbp]
\centering
\small
\setlength{\tabcolsep}{4pt}
\renewcommand{\arraystretch}{1.1}
\begin{tabular}{lc}
\toprule
Item & Value \\
\midrule
Max search calls & 1 \\
Retrieved documents & 5 \\
Temperature & 0.7 \\
Top-$p$ & 0.9 \\
Repetition penalty & 1.1 \\
Max generation length & 1024 \\
\bottomrule
\end{tabular}
\caption{Inference hyperparameters.}
\label{tab:repro_inference_config}
\end{table}

\subsection{Baseline Implementation}

\label{baseline_details}

We implement or adapt all baselines under the same model scale, retrieval setup, and evaluation protocol whenever applicable.
For retrieval-based baselines, the same retrieved documents are used.
For critique- or refinement-based baselines, the same critic backbone is used unless otherwise specified.

\begin{itemize}[leftmargin=0.5cm]
    \item \textbf{Vanilla}: The base language model without retrieval, directly generating answers from the question.
    \item \textbf{Naive RAG~\cite{lewis2020retrieval}}: Retrieved documents are directly concatenated with the question as input, without additional prompt engineering.
    \item \textbf{CoT~\cite{wei2022chain}}: Augments the generation process with intermediate reasoning steps, enabling the model to perform multi-step reasoning before producing the final answer.
    \item \textbf{Self-RAG~\cite{asai2023self}}: Introduces self-reflection and iterative retrieval to refine intermediate reasoning steps during generation.
    \item \textbf{Self-Refine~\cite{madaan2023self}}: Performs iterative answer refinement by generating feedback and improving responses over multiple rounds.
    \item \textbf{Align-RAG~\cite{weiretrieval}}: Enhances RAG through alignment-based optimization to improve consistency between retrieved evidence and generated answers.
    \item \textbf{Search-R1~\cite{jin2025search}}: A strong baseline that integrates structured search and reasoning trajectories for improved retrieval and generation.
    \item \textbf{\methodname}: Our proposed method, which introduces an external critic to provide structured feedback for improving answer correctness and reasoning quality.
\end{itemize}

\paragraph{Implementation Details.}
To ensure a fair and reproducible comparison, we use a unified experimental setup across all methods.
All generation-based methods use Qwen2.5-7B-Instruct as the backbone model, and all critique- or feedback-based methods use Qwen2.5-3B-Instruct as the critic model.
For retrieval-based methods, we use the same Wikipedia-based corpus, retriever, and index, with a single retrieval call and top-$5$ retrieved documents for each question.
We do not tune retrieval depth or retrieval configuration separately for different datasets.
For baselines that require critique or refinement, such as Align-RAG, we implement them under the same model scale, retrieval setting, prompt format, and inference pipeline as our method whenever applicable.
Finally, all methods are evaluated using the same decoding configuration, answer extraction procedure, evaluation metrics, and LLM-based judging protocol.

\subsection{Hardware and 
Environment}
\label{app:hardware_software}

Critic training is conducted on NVIDIA A100 40GB GPUs, while inference and evaluation are conducted on NVIDIA V100 32GB GPUs.

\begin{table}[htbp]
\centering
\small
\begin{tabular}{ll}
\toprule
Item & Version / Configuration \\
\midrule
Python & 3.12.0 \\
PyTorch & 2.8.0+cu128 \\
CUDA & 12.8 \\
Transformers & 4.56.1 \\
vLLM & 0.15.1 \\
FAISS & 1.8.0 \\
\bottomrule
\end{tabular}
\caption{Main software dependencies.}
\label{tab:software_env}
\end{table}

\subsection{Dataset Statistics}

\begin{table}[h]
\centering
\resizebox{\columnwidth}{!}{
\begin{tabular}{lccc}
\toprule
Dataset & Split & Usage & \#Examples \\
\midrule
HotpotQA & Train & Critic training & 5,000/90447 \\
HotpotQA & Dev & In-domain & 7,405/7,405 \\
NQ & Dev & Out-of-domain & 3,610/3,610\\
TriviaQA & Validation & Out-of-domain & 17,944/17,944 \\
ASQA & Dev & Out-of-domain & 948/948 \\
PopQA & Test & Out-of-domain & 14,267/14,267 \\
\bottomrule
\end{tabular}
}
\caption{Dataset statistics used in our experiments.}
\label{tab:dataset_statistics}
\end{table}

\section{Additional Experimental Results}
\label{app:additional_results}
\subsection{Detailed Per-Run Results for Efficiency Comparison}
\label{app:rq3_details}

To complement the averaged results reported in Table~\ref{tab:rq3_efficiency}, 
we provide detailed per-run results for the efficiency comparison in this appendix.

For Continued Generator Training, results are reported from two independent inference runs on both HotpotQA and NQ, and the main-text results are obtained by averaging these runs.
For Generator + Trained Critic, we report the corresponding inference results used in the main table.

These detailed results help illustrate the stability of continued generator training and provide additional transparency for the efficiency comparison under equal supervision budget.

\begin{table}[htbp]
\centering
\small

\label{tab:rq3_hotpot_runs}
\setlength{\tabcolsep}{4pt}
\renewcommand{\arraystretch}{1.15}
\resizebox{\columnwidth}{!}{
\begin{tabular}{c|ccc}
\toprule
Method & F1 & SBERT & Acc \\
\midrule

Continued Generator Training (Run 1)
& 52.8 & 67.7 & 51.9 \\

Continued Generator Training (Run 2)
& 46.4 & 62.1 & 51.4 \\

Average
& 49.6 & 64.9 & 51.7 \\

\midrule

Generator + Trained Critic
& 52.2 & 71.0 & 57.5 \\

\bottomrule
\end{tabular}
}
\caption{
Detailed per-run results for efficiency comparison on NQ.
}
\end{table}

\begin{table}[htbp]
\centering
\small

\label{tab:rq3_nq_runs}
\setlength{\tabcolsep}{4pt}
\renewcommand{\arraystretch}{1.15}
\resizebox{\columnwidth}{!}{
\begin{tabular}{c|ccc}
\toprule
Method & F1 & SBERT & Acc \\
\midrule

Continued Generator Training (Run 1)
& 41.4 & 62.4 & 49.6 \\

Continued Generator Training (Run 2)
& 48.8 & 64.8 & 51.4 \\

Average
& 45.1 & 63.6 & 50.5 \\

\midrule

Generator + Trained Critic
& 46.3 & 65.4 & 50.6 \\

\bottomrule
\end{tabular}
}
\caption{
Detailed per-run results for efficiency comparison on HotpotQA.
}
\end{table}

\subsection{Case Studies for \methodname}
\label{appendix:case_studies}

We provide representative examples to illustrate how \methodname~supports refinement.
Rather than presenting full trajectories, we summarize the key evidence, critique signal, and refinement outcome for readability.

\subsection{Case A: Critic-Guided Error Correction}

The following example illustrates how CRITIC-R1 identifies an error in the initial 
trajectory and provides critique that leads to a corrected version.
\begin{figure}[htbp]
    \centering
    \includegraphics[width=1\linewidth]{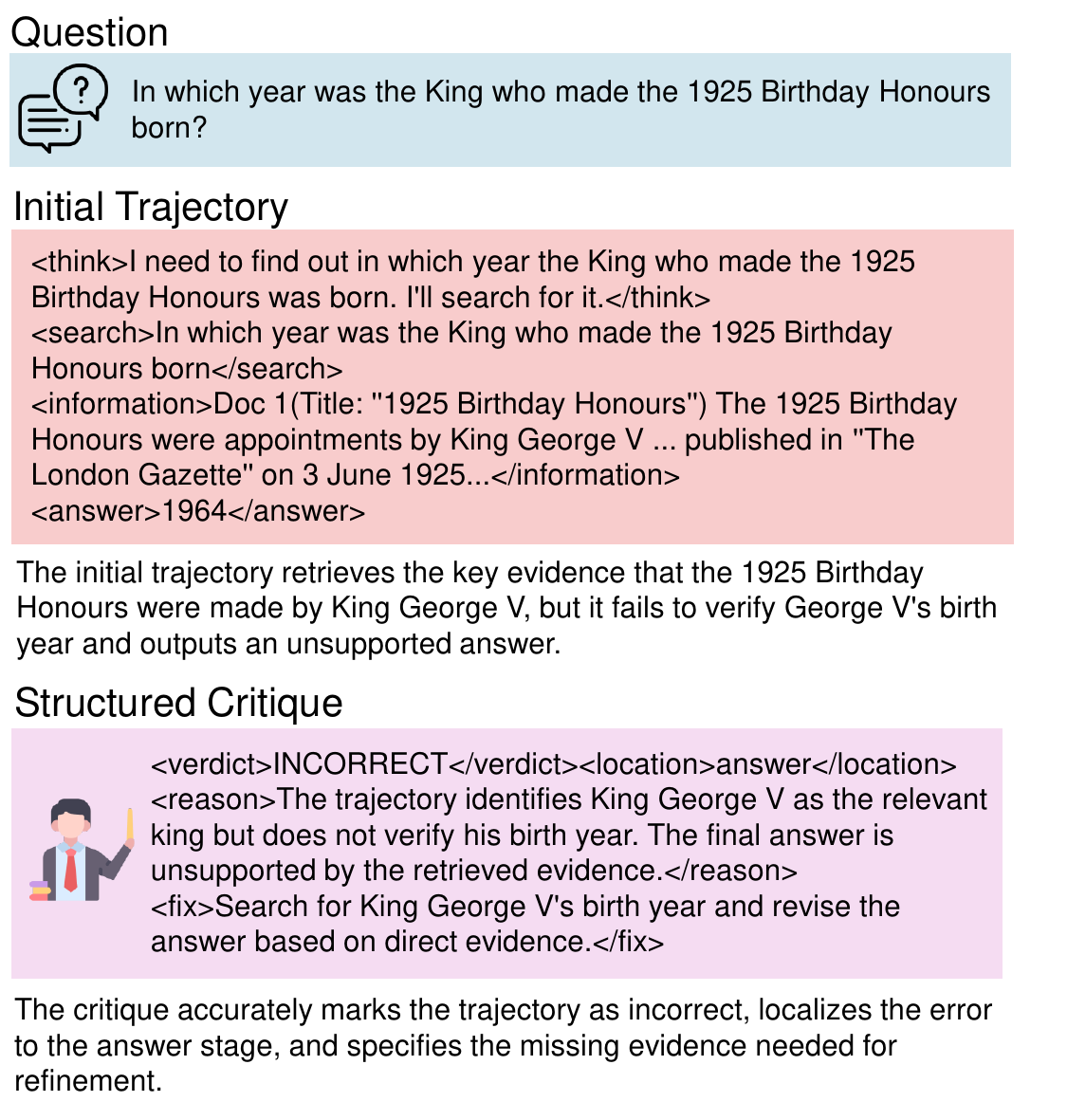}
\end{figure}

\vspace{2mm}

\begin{figure}[htbp]
    \centering
    \includegraphics[width=1\linewidth]{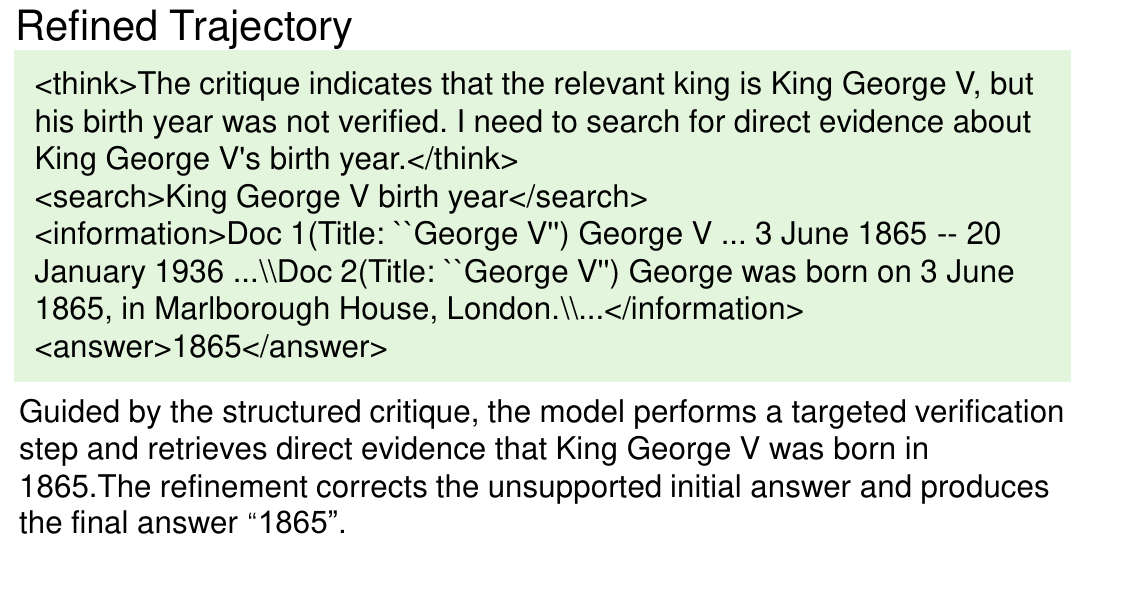}
\end{figure}

\subsection{Case B: Selective Non-Intervention}

The following example illustrates how CRITIC-R1 recognizes that the initial 
trajectory has no obvious error and therefore chooses not to intervene, 
thereby avoiding over-aggressive intervention on an already correct answer.

\begin{figure}[htbp]
    \centering
    \includegraphics[width=1\linewidth]{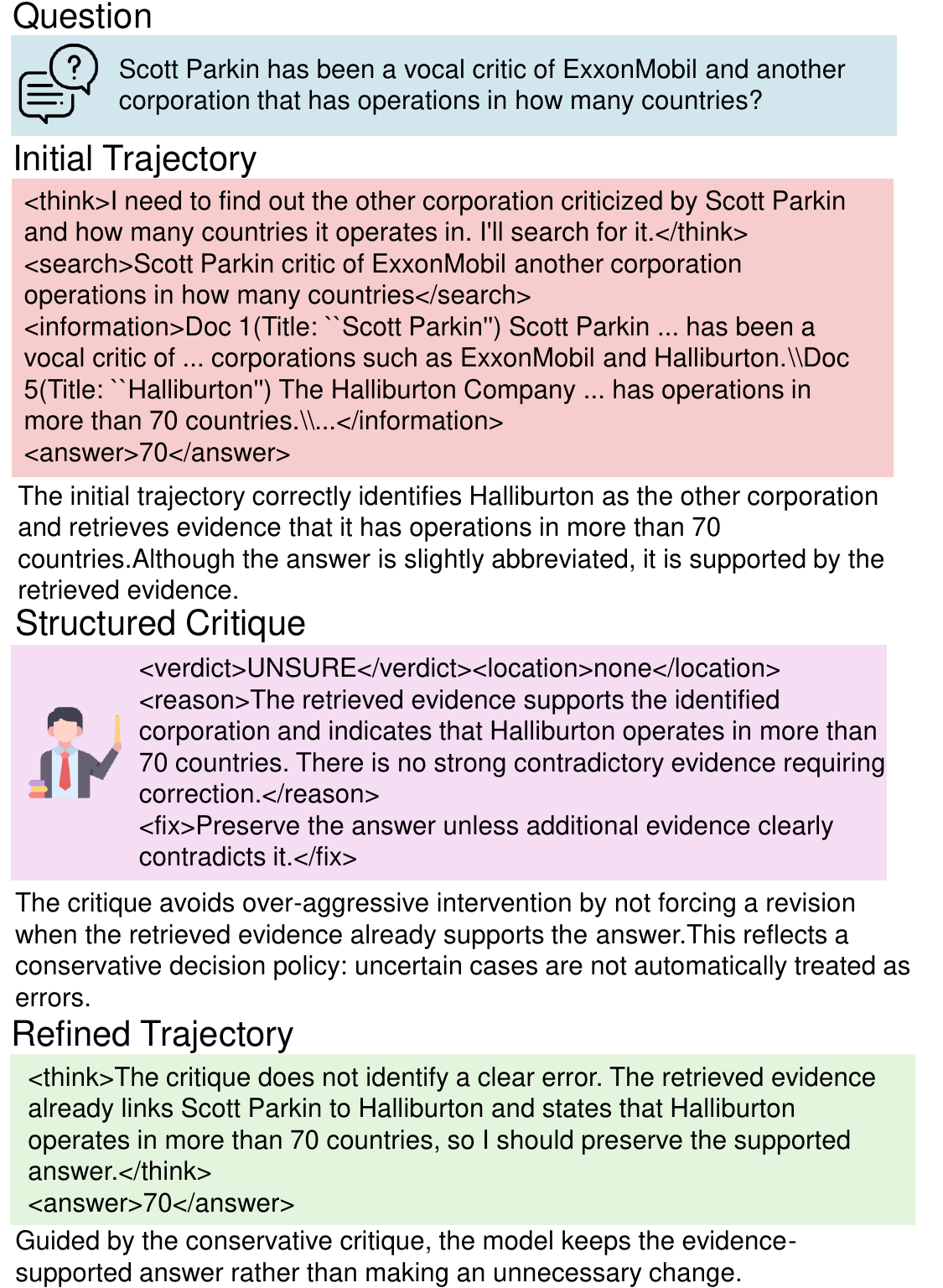}
\end{figure}

\subsection{Case C: Critique-Guided Evidence Selection Repair}

The following example illustrates how CRITIC-R1 identifies an evidence-selection error in the initial trajectory and provides critique that guides the model to select the evidence matching the question constraint.

\begin{figure}[htbp]
    \centering
    \includegraphics[width=1\linewidth]{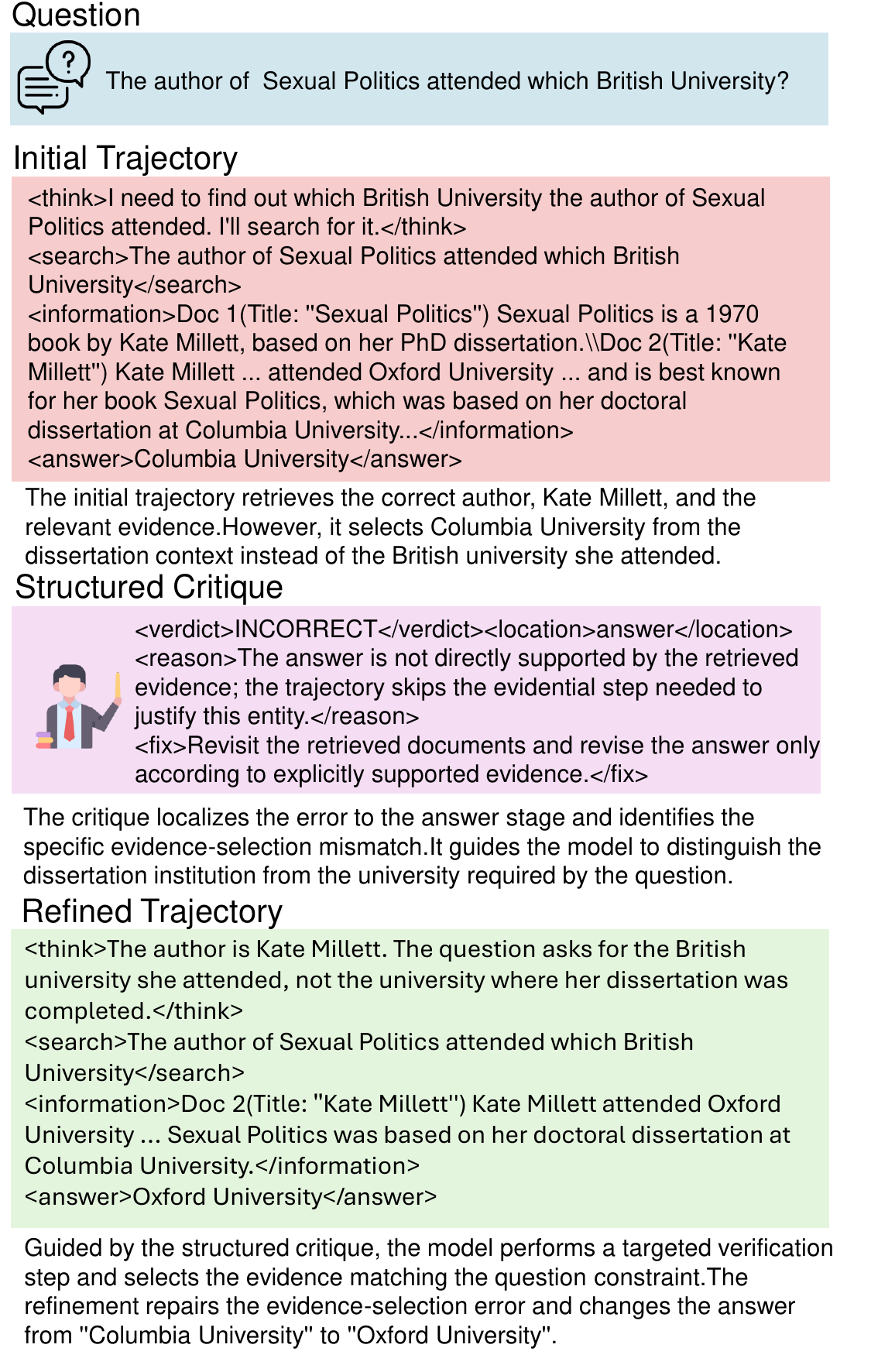}
\end{figure}

\section{Additional Discussions}
\label{app:additional_discussions}

\subsection{Artifacts and Licenses}
\label{app:artifacts_licenses}

We will release our code, prompts, and evaluation scripts for research purposes. 
All datasets used in this work are publicly available, and their use follows the corresponding dataset licenses and terms. 
Our use of these datasets and pretrained models is limited to research-oriented training, evaluation, and analysis, which is consistent with their intended research use. 
Any derived supervision data or model artifacts will be released only when permitted by the licenses and terms of the original datasets and base models.

\subsection{Potential Risks}
\label{app:potential_risks}

Our critic may still produce incorrect diagnoses, especially when retrieved evidence is incomplete or misleading. 
Since the supervision relies on LLM-as-judge annotations, it may also inherit biases or errors from the judge model. 
The method should therefore not be treated as a guarantee of factual correctness, and human verification remains necessary in high-stakes applications.

\subsection{Data Privacy and Content}
\label{app:data_privacy}

We use publicly available QA benchmarks and do not collect new data from human participants or private user interactions.  
We do not intentionally include personally identifying information, and any released artifacts will exclude private metadata, local paths, API keys, or other identifiers. 

\subsection{Use of AI Assistants}
\label{app:ai_assistants}

We used AI assistants to assist with language polishing, grammar checking, code debugging, and improving clarity of presentation.

\end{document}